\documentclass[letterpaper, 10 pt, conference]{ieeeconf} 
\usepackage[utf8x]{inputenc}
\usepackage[dvipsnames]{xcolor}
\usepackage{algorithm}
\usepackage[noend]{algpseudocode}
\usepackage{multicol}
\usepackage[subrefformat=simple,labelformat=simple]{subcaption}

\IEEEoverridecommandlockouts                              % This command is only
\definecolor{green}{RGB}{27,158,119}
\definecolor{darkgreen}{RGB}{7,98,19}
\definecolor{red}{RGB}{240,50,2}
\definecolor{blue}{RGB}{50,65,190}

\usepackage{tikz}
\usepackage{pgfplots}
\usepackage{amsmath}
\usepackage{amsfonts}
\usepackage{amssymb}
\usepackage{tkz-graph}
\usepackage{dirtytalk}
\usetikzlibrary{calc,arrows,plotmarks,patterns}

\newcommand\xaxis{190}
\newcommand\yaxis{-45}
\newcommand\zaxis{90}

\newcommand\topside[4]{
  \fill[fill=black!#4, draw=black,shift={(\xaxis:#1)},shift={(\yaxis:#2*0.7)},
  shift={(\zaxis:#3)}] (0,0) -- (10:1) -- ++(135:0.7) --++(190:1)--(0,0);
}

\newcommand\leftside[4]{
  \fill[fill=black!#4, draw=black,shift={(\xaxis:#1)},shift={(\yaxis:#2*0.7)},
  shift={(\zaxis:#3)}] (0,0) -- (0,-1) -- ++(135:0.7) -- ++(0,1)--(0,0);
}

\newcommand\rightside[4]{
  \fill[fill=black!#4, draw=black,shift={(\xaxis:#1)},shift={(\yaxis:#2*0.7)},
  shift={(\zaxis:#3)}] (0,0) -- (10:1) -- ++(0,-1) --(0,-1)--(0,0);
}

\newcommand\cube[6]{
  \topside{#1}{#2}{#3}{#4} \leftside{#1}{#2}{#3}{#5} \rightside{#1}{#2}{#3}{#6}
}

\newcommand\topsidea[3]{
  \fill[fill=blue!#30, draw=black,shift={(\xaxis:#1)},shift={(\yaxis:#2*0.7)},
  shift={(\zaxis:#3)}] (0,0) -- (10:1) -- ++(135:0.7) --++(190:1)--(0,0);
}

\newcommand\leftsidea[3]{
  \fill[fill=blue!60, draw=black,shift={(\xaxis:#1)},shift={(\yaxis:#2*0.7)},
  shift={(\zaxis:#3)}] (0,0) -- (0,-1) -- ++(135:0.7) -- ++(0,1)--(0,0);
}

\newcommand\rightsidea[3]{
  \fill[fill=blue!80, draw=black,shift={(\xaxis:#1)},shift={(\yaxis:#2*0.7)},
  shift={(\zaxis:#3)}] (0,0) -- (10:1) -- ++(0,-1) --(0,-1)--(0,0);
}

\newcommand\cubeAgent[3]{
  \topsidea{#1}{#2}{#3} \leftsidea{#1}{#2}{#3} \rightsidea{#1}{#2}{#3}
}

\newcommand\topsidel[3]{
  \fill[fill=red!80, draw=black,shift={(\xaxis:#1)},shift={(\yaxis:#2*0.7)},
  shift={(\zaxis:#3)}] (0,0) -- (10:1) -- ++(135:0.7) --++(190:1)--(0,0);
  }

\newcommand\leftsidel[3]{
  \fill[fill=red!90, draw=black,shift={(\xaxis:#1)},shift={(\yaxis:#2*0.7)},
  shift={(\zaxis:#3)}] (0,0) -- (0,-1) -- ++(135:0.7) -- ++(0,1)--(0,0);
}

\newcommand\rightsidel[3]{
  \fill[fill=red!95, draw=black,shift={(\xaxis:#1)},shift={(\yaxis:#2*0.7)},
  shift={(\zaxis:#3)}] (0,0) -- (10:1) -- ++(0,-1) --(0,-1)--(0,0);
}

\newcommand\cubeLava[3]{
  \topsidel{#1}{#2}{#3} \leftsidel{#1}{#2}{#3} \rightsidel{#1}{#2}{#3}
}

\title{\LARGE \bf
Accelerating Empowerment Computation with UCT Tree Search
}

\author{Christoph Salge$^{1,2}$, Christian Guckelsberger$^{3}$, Rodrigo Canaan$^{2}$ and Tobias Mahlmann
\thanks{$^{1}$School of Computer Science, University of Hertfordshire, College Lane, AL10 9AB, Hatfield, UK {\tt\small c.salge@herts.ac.uk}}
\thanks{$^{2}$Department of Computer Science and Engineering, New York University, 5 MetroTech Center, 11201, Brooklyn, NY, USA {\tt\small rodrigo.canaan@nyu.edu}}
\thanks{$^{3}$Computational Creativity Group Goldsmiths, University of London
London, SE14 6NW, UK {\tt\small c.guckelsberger@gold.ac.uk}}
}

\begin{document}

\maketitle
\thispagestyle{empty}
\pagestyle{empty}

%%%%%%%%%%%%%%%%%%%%%%%%%%%%%%%%%%%%%%%%%%%%%%%%%%%%%%%%%%%%%%%%%%%%%%%%%%%%%%%%

\begin{abstract}
Models of intrinsic motivation present an important means to produce sensible behaviour in the absence of extrinsic rewards. Applications in video games are varied, and range from intrinsically motivated general game-playing agents to non-player characters such as companions and enemies. The information-theoretic quantity of \emph{Empowerment} is a particularly promising candidate motivation to produce believable, generic and robust behaviour. However, while it can be used in the absence of external reward functions that would need to be crafted and learned, empowerment is computationally expensive. In this paper, we propose a modified UCT tree search method to mitigate empowerment's computational complexity in discrete and deterministic scenarios. We demonstrate how to modify a \emph{Monte-Carlo Search Tree with UCT} to realise empowerment maximisation, and discuss three additional modifications that facilitate better sampling. We evaluate the approach both quantitatively, by analysing how close our approach gets to the baseline of exhaustive empowerment computation with varying amounts of computational resources, and qualitatively, by analysing the resulting behaviour in a \emph{Minecraft}-like scenario. 
\end{abstract}

\section{Introduction}

The empowerment formalism \cite{Klyubin2008,Salge2014} offers interesting game applications in terms of believable NPC behaviour \cite{guckelsberger2016intrinsically}, general game-play \cite{Anthony2014} and player experience modelling \cite{guckelsberger2017predicting}. But the high computational complexity of empowerment is problematic for a wider application in games. In this paper we show how a UCT tree search formalism \cite{kocsis2006bandit,browne2012survey} can be adapted to approximate empowerment maximisation in the discrete domain. But first, we motivate the application of empowerment to games in more detail. 

\subsection{Motivation}

Empowerment is a measure of how much an agent can affect the world it itself perceives. Empowerment maximisation is considered an intrinsic motivation (IM)  \cite{Oudeyer2008}, and has been recently linked to competence and autonomy, two motivations which are frequently discussed in a games context \cite{Roohi2018}. As intrinsic motivation, i.e. as an essential motivation linked to agency itself, empowerment can generate behaviour even in the absence of externally defined goals. Behaviour then results from fulfilling a motivation that arises from the agent-world interaction. An illustrative, non-empowerment example for this is the work of Merrick and Maher \cite{Merrick2008,Merrick2009}, where an agent's actions are selected based on learning progress~\cite{4141061} and curiosity~\cite{Schmidhuber2006}. Curiosity, or the desire to experience something new, can create behaviour without further reward. The broad concept of curiosity is also a good illustration of an \textit{intrinsic} motivation, as it is hard to imagine agency without the least desire to experience novelty or learn something new. Empowerment, in contrast, is about having affordances, about self-efficacy and the ability to affect one's own world. Empowerment has also been linked to the idea of an organism's striving to preserve its precarious existence \cite{Guckelsberger2016}. Applied to a Minecraft-like simulation this drive to self-preservation resulted in behaviour where the agent would restructure the world to keep itself alive \cite{Salge2014a}, producing different behaviour patterns in reaction to changes in the environment. The same work also demonstrated how the embodiment of the agent was reflected in the structures built in the world. This apparent self-directed behaviour, arising from- and reacting to changes in the game world make the application of empowerment in games interesting.

Previously, empowerment has been applied to play Sokoban and PacMan \cite{Anthony2014}. In this work, Anthony et al. speak about the generality of empowerment by asserting that it provides a utility that: \say{1) derives
only from the structure of the problem itself and not from an
external reward; 2) identifies the desirability of states in a way
that matches intuition; and 3) carries over between scenarios of
apparently different character.} This would make empowerment a good proxy for general game-play, and thus biasing the decision making of reward-optimising agents with empowerment might lead to better performance. Similar approaches have been used in the domain of robotics, where a robotic follower \cite{leu2013empRobot} and underwater vehicles \cite{abreu2016widely} had their decision making biased or enhanced with empowerment maximisation. But while the empowerment formalism is generally applicable, i.e. can be computed based just on the structure of a given forward model, the utility it provides may not always agree with an externally defined reward. Hence, it is possible to define a (possibly contrived) win condition that conflicts with empowerment. In a lot of cases, though, games are designed to be aligned with intrinsic motivations, and progressing in a game usually goes along with increasing player empowerment.

Creating believable non-player characters in a game is another possible application of empowerment in games. Empowerment maximisation can, without defined goals, produce behaviour related to self-preservation and maximisation of options. This can be used to give NPCs an appearance of self-determination. The general applicability of the formalism also allows NPCs to adapt to changing circumstances. This has been explored by Guckelsberger et.al. \cite{guckelsberger2016intrinsically} for the design of general companion characters. In addition to having the companion NPC maximise their own empowerment, two additional empowerment drives where introduced. Maximising the empowerment of the player motivates the companion to protect and help. It would, for example, shoot enemies that try to kill the player. This multi-perspective approach has also been explored in relation to robots, where it could produce generic robot behaviour guidelines \cite{salge3laws}.

Finally, there is also the question to which extent a player's empowerment in a game can be used as predictor for their experience. A preliminary study \cite{guckelsberger2017predicting} identified causal efficacy as potential candidate experience that empowerment is closely related to, with mediate effects on \say{challenge, involvement, attention and engagement, learning and emotions}. Having a measure that computes user experience without an actual player would be beneficial in rapid prototyping and when creating or adapting games automatically.

Yet, one downside of empowerment maximisation is its lack of scalability, due to the formalism's high computational complexity, especially when looking at longer time horizons. Approximations have been developed both for the \emph{continuous} domain \cite{salge2012approximation} and discrete but \emph{noisy} models \cite{Anthony2014}. In this paper, we use UCT to accelerate the computation of the most empowered action in a \emph{discrete and deterministic} model. %%In this paper we accelerate finding the most empowered action in a discrete and deterministic model. 

\subsection{Overview}

We first describe the actual empowerment formalism, and then focus on empowerment in discrete and deterministic models. We discuss the problems arising from sparse sampling, and introduce a modified UCT tree search algorithm to find the most empowered actions with less sampling. We complement this with three modifications -- \emph{novelty bias}, \emph{aggregated empowerment} and \emph{full branching} -- to further enhance the sampling. We evaluate the optimisation scheme in a Minecraft-like world model which has also been used in previous work \cite{Salge2014a}. We briefly introduce the model, followed by a quantitative and a qualitative evaluation. We demonstrate how the UCT approach and the different modifications perform better with less samples than the current baseline of sparse random sampling. 

%\section{Background}
\section{Empowerment}
\label{sec:empowerment}

\begin{figure}[thb]
\begin{center}
\resizebox{1.0\linewidth}{!}{
\begin{tikzpicture}[scale=0.7]
  \GraphInit[vstyle=Dijkstra]
  \SetVertexMath \renewcommand*{\VertexLineColor}{white}
  
  \SetGraphUnit{1.5}
  \Vertex[L=R_{t+1},Lpos=90]{R0}
  \SOEA[L=S_{t+1},Lpos=-90](R0){S0}
  \EA[L=A_{t+1},Lpos=-90](S0){A0}
  \NOEA[L=R_{t+2},Lpos=-90](A0){R1}
  \SOEA[L=S_{t+2},Lpos=-90](R1){S1}
  \EA[L=A_{t+2},Lpos=-90](S1){A1}
  \NOEA[L=R_{t+3},Lpos=-90](A1){R2}
	
  \WE[empty](R0){E0}
  \SOWE[L=A_{t}](R0){E1}
  \EA[empty](R2){E2}
  \SOEA[L=S_{t+3}](R2){E3}
  	
  \SetUpEdge[style={post}]
  \tikzstyle{EdgeStyle}=[line width=.8pt]
  \tikzstyle{LabelStyle}=[left=3pt]
  \tikzstyle{LabelStyle}=[above=3pt]
  \Edge(R0)(R1)
  \Edge(R1)(R2)
  \Edge(R0)(S0)
  \Edge(S0)(A0)
  \Edge(A0)(R1)
  \Edge(R1)(S1)
  \Edge(S1)(A1)
  \Edge(A1)(R2)
  \Edge(R2)(E3)
  \Edge(E1)(R0)
	
  \SetUpEdge[style={post}, color=gray]
  \tikzstyle{EdgeStyle} = [line width=.8pt]
  \Edge(E0)(R0)

  \Edge(R2)(E2)

  \SetUpEdge[style={post,dotted,color=red,bend right = 20}]
  \Edge(E1)(E3)
  \Edge(A0)(E3)
  \Edge(A1)(E3)
\end{tikzpicture}
}
\end{center}
\textbf{\refstepcounter{figure}\label{fig:pal1} Figure \arabic{figure}.}{ The perception-action-loop visualised as a Bayesian network. \textit{S} is the sensor, \textit{A} is the actuator, and \textit{R} represents the rest of the system. The index $t$ indicates the time at which the variable is considered. This model is a minimal model for a simple memoryless agent. The red arrows indicate the direction of the potential causal flow relevant for 3-step empowerment.}
\end{figure} 

\section{Empowerment}

Empowerment \cite{Klyubin2008,Salge2014a} is an information-theoretic formalism that captures how much an agent can affect the world it itself perceives. It is defined for all systems that can be modelled as an action-perception loop, as seen in Fig.~\ref{fig:pal1}. Where the random variables $S$, $A$ and $R$ model the sensors, actions and remaining state of the world, respectively. Empowerment for a given state $r \in R$ is formally defined as the channel capacity from an agent's actions at time $t$ to its sensors at a later point in time. This channel goes through the environment $R$. A common generalisation is $n$-step empowerment, where all actions from $a_t$ to $a_{t+n-1}$ are considered as input to the channel, and the output is the sensor of the agent at $t+n$:
\begin{equation}
\label{channelCapacity}
\mathfrak{E}(r_t) = \max_{p(a_{t...t+n-1})} I(S_{t+n};A_{t...t+n-1}|r_t).
\end{equation}
The quantity captures how much information an agent can ``inject'' into its sensor $S_{t+n}$ via the environment by intervening earlier in $A_{t...t+n-1}$. It is equivalent to potential causal information flow as defined in \cite{ay2008information}. An agent is usually high empowered if it has a lot of different options that all lead to different, predictable outcomes, unaffected by noise. A highly empowered agent can reliably bring about many different sensor states. A more detailed discussion of the general concept and its information theoretic basis can be found in \cite{Klyubin2008,Salge2014a}.  

Empowerment maximisation is the idea that an agent wants to be in a state that is highly empowered. Note, when computing the empowerment for a given state the channel capacity achieving distribution $p(a_{t...t+n-1})$ might contain a lot of action sequences that lead to bad outcomes. But the action policy, i.e the way the agent picks it actions, is not determined by this distribution. Instead, a greedy empowerment maximisation strategy computes the empowerment for all possible successor states to the current states, and then chooses the action leading to the one with the most empowerment. It is empowerment maximisation that is considered an intrinsic motivation, and in this paper, we are focussing on how to efficiently determine which actions will lead us to the most empowered state.

While empowerment can be defined for both a noisy and even a continuous channel, in this paper we focus on discrete and deterministic models. In the deterministic case, where each possible action sequence $a_t, ..., a_{t+n-1}$ leads to one specific state $s_{t+n}$, the channel capacity is the logarithm of all reachable states. So, to calculate the empowerment we have to determine the resulting sensor state for each possible n-step action sequence, and then count how many different states there are in total. This simplifies the computation significantly, and allows to compute n-step empowerment for larger time horizons $n$. We will refer to these reachable sensor states as ``reachable states'', dropping the word sensor for brevity.  
Despite this simplification, if we look at a model where each action step has, for example, a branching factor of $5$, we still have to evaluate $5^n$ action sequences and the corresponding final states. This number quickly grows infeasibly. In previous work \cite{Salge2014a} this was addressed with sub-sampling, where a random subset of all possible action sequences was evaluated to compute $15$-step empowerment, with a branching factor of $12$. In this work the limitations of random sampling became evident. Sometimes, the empowerment-maximising agent would be in a situation where it could get to a part of the world where it would be able to reach a lot of different sensor states, but to get there it would have to perform very specific actions in the beginning of the action sequence - think of a bridge as an evocative example. Due to the random nature of the sampling, this bridge might only be crossed with a few of the sequences, and the evaluation would miss the considerable gain that going over the bridge yields for the agent's empowerment. In this work, we aim to use UCT tree search to both (i) bias the exploration towards those initial sequences, and to (ii) identify the best possible action more efficiently. 

\section{Empowerment with UCT Tree Search}

In this section we outline how to use tree search with UCT (upper confidence bound applied to tree search, \cite{kocsis2006bandit}) to accelerate the decision making based on deterministic empowerment. To find the best action, we need to determine which of the successor states of the current world state has the most different sensor states reachable with action sequences of length $n$. Note, that while we compute the empowerment for sensor states derived form the world, the computations to determine the empowerment are done with a complete model of all world state transitions. In other words, our computation is not limited by the agent's perspective.

Our approach is inspired by Monte Carlo tree search (MCTS) with UCT \cite{browne2012survey}, but there are substantial adaptations in the expansion, simulation and backpropagation steps. The basic idea of MCTS UCT, or any informed search for that matter, is to guide the use of resources, such as forward model calls, to the parts of the search space which yield the most information for picking the best action.
So, when we sample action sequences to determine reachable sensor states, we assume that sequences starting with actions that have led to new results are more likely to yield new results again. We then use the UCT formula \cite{kocsis2006bandit} to bias our exploration towards those actions that have previously led to new states. Further analysis is still needed to determine if the mathematical properties of the bandit problem, which UCT is derived from, hold for empowerment computation. Here we look at simulated results only.  

In the next section, we describe the algorithm in detail, and motivate three modifications. Both are further illustrated with pseudocode in Alg.~1. Keep in mind, the goals is to determine which successor state of the current world state has the most reachable sensor states, i.e. is the most empowered. 

\begin{algorithm*}[htb]

\caption{Overview of the agent's decision making algorithm. Methods that take place in other objects, such as applying actions to the world or selecting a random number are omitted for brevity. HORIZON is $n+1$ for $n$-step empowerment, as we can determine the initial successor states in the same tree. The colours indicate code changes for the modification. For basic UCT empowerment read just the code in black. For \textit{aggregated empowerment}, add the \textcolor{red}{red} line at 18. For \textit{novelty bias} add the two \textcolor{blue}{blue} code snippets in 29 and 36. For the \textit{full branching} modification add the code in \textcolor{green}{green}, and set DEPTH to a non-zero value.}
\label{algo:uct}
\begin{multicols}{2}
\begin{algorithmic}[1]
\linespread{1}\selectfont
\Procedure{Best\_Action}{World $w$}
	\State Node $root \gets$ new Node

    \While {time left}
   	    \State $depth \gets 0$
	    \State Node $t \gets root$
	    \State World $test \gets $copy($w$)
		\While {$depth <$ HORIZON \textcolor{green}{- DEPTH}}
        	\State $t$.visits++
            \State $depth$++
            \If{$t$ has unexpanded children} 
				\State Action $a \gets$ \textproc{Random\_Action}($t$, $test$)
				\State $test$.applyAction($a$)
                \State Node $child \gets$ new Node
				\State $child$.parent $\gets t$        				
                \State $child$.action $\gets a$
                \State $t$.children.add($child$)
               	\State $state \gets test$.s() \Comment Agent sensor state
                \textcolor{red}{\State \textproc{Add\_State}($state$, $child$)}
                \State $t \gets child$
			\Else \Comment use UCT selection
            	\State $t \gets$ \textproc{Best\_Child}($t$, $root$)
                \State $test$.applyAction($t$.action)
            \EndIf
        \EndWhile
        \State \textproc{Branch}($t$, $test$, DEPTH, $root$)
    \EndWhile
    \State \Return $root$.children[max(states)].action
\EndProcedure
\State
\Procedure{Add\_State}{$state$, Node $n$}
    \If{ $state$ $\not\in$ $n$.states }
    	\State $n$.states.add($state$)
        \color{blue}\If{$state \not\in parent$.states} $n$.unique++
        \EndIf
        \color{black}\If{$\exists$ $n$.parent} \textproc{Add\_State}($state$, $n$.parent)
        \EndIf
    \EndIf    
\EndProcedure
\State
\Procedure{Best\_Child}{Node $t$, Node $root$}
	\State $best \gets$ null
    \State $fitness \gets$ 0
	\For{$c \in t$.children} 
    	\State $f$=\large$\frac{|c.states|+\textcolor{blue}{c.unique}}{c.visits} + 0.01\cdot\sqrt[]{\frac{\log root.visits}{c.visits}}$\normalsize
        \linespread{2}\selectfont
    	\If{$f > fitness$}
        	$fitness \gets f$;
        	$best \gets c$
        \EndIf
        \linespread{1}\selectfont
    \EndFor
	\State
    \Return $best$
\EndProcedure
\State
\Procedure{Branch}{Node $t$, World $w$, $d$, Node $root$}\color{green}
\If{$d$ = 0} \color{black}
    	\State $state \gets test$.s() \Comment Agent sensor state
        \State \textproc{Add\_State}($state$, $t$, $root$) \color{green}
    \Else
        \State Action[] $a \gets w$.getPossibleActions()
		\For{$action \in a$}
   	 		\State World test $\gets$ copy($w$)	
			\State $test$.applyAction($action$)            
            \State Node $child \gets$ new Node
			\State $child$.parent $\gets t$        				
            \State $child$.action $\gets action$
            \State $t$.children.add($child$)
            \State \textproc{Branch}($child$, $test$, $d$ - 1, $root$)
        \EndFor
	\EndIf \color{black}
\EndProcedure
\State
\Procedure{Random\_Action}{Node $t$, World $w$}
	\State Action[] $a \gets w$.getPossibleActions()
	\For {actions $\in$ $t$.children}
    	$a$.remove(action)
    \EndFor
    \State
	\Return random($a$)
\EndProcedure
\end{algorithmic}
\end{multicols}
\end{algorithm*}

\subsection{UCT tree search}

In our algorithm nodes are associated with world states. We start by creating a root node that is associated with the world in its current state. This node has a depth of zero. 

\subsubsection{Expansion} The algorithm starts at the root, and checks if there are unexpanded, i.e. unvisited children. As long as there are unvisited children we select randomly one of the actions that would lead to an unexpanded child. We create the child node with a depth value one higher than its parent node. We then repeat the expansion step, i.e. expanding another unvisited child, until we reach a node that has a depth equal to the empowerment horizon $n$ plus one. This is because the first step just creates the successor nodes that we are evaluation for their empowerment, and the successive $n$ steps realise the $n$-step empowerment approximation. 

\subsubsection{Backpropagation} Once we reach a node at the horizon, we obtain the agent's sensor state and store it in the set of reachable states in the node. We then also add this reachable state to the node's parent node. The parent checks if it already has this state in its reachability set, and if not, adds it. It then also adds it to its own parent, recursively. After the backpropagation finishes, the tree should be in a state where each node has a reachability set that contains all sensor states that can be reached from it with the already expanded action sequences. After this step the algorithm starts over at the root of the tree, if there is time left. 

\subsubsection{Selection}
As the tree fills up, the algorithm will encounter nodes where all children have at least been visited once, and it then has to decide which node to expand again. At this point we sort the children $c$ with the modified UCT formula and pick the child $c$ with the highest value:

\begin{equation}
uct(c) = \frac{c.states.size()} {c.visits} + 0.01 \cdot \sqrt[]{\frac{\log{(root.visits)}}{c.visits}}
\end{equation}

The function $states.size()$ gives us the size of the reachability set, and $visits()$ tells us how often the root node and the child node have been visited. The size of the reachability set divided by the visits to the child gives us a value between $0.0$ and $1.0$, the ratio of how many new states each visit to this particular child found. The best value here is $1.0$, meaning every visit leads to one new sensor state. The term in the square root guides selection towards under-explored states; it grows larger as more action sequences are sampled that do not go through $c$. The factor of $0.01$ was determined experimentally, as it lead to a good balance where the distribution of visits among children was neither approximately uniform (as it would be for a larger value), nor heavily skewed towards the first best solution (as it would be for a smaller value). 

\subsubsection{Action Selection}

At some point the algorithm runs out of time (or some other computational limiter) and it needs to decide what action to take. It looks at the root and picks the action leading to the child with the biggest reachability set. This action should lead to a state with the highest $n$-step empowerment, meaning that from this state, the agent can reach the maximum number of different sensor states within $n$ steps. Note that the algorithm does not select the state with the highest ratio of new states vs. visits, which was used for the UCT based selection. 

\textbf{Rationale:} Compared with randomly sampling action sequences for each successor state, this algorithm should perform better at finding the state with the highest empowerment, as it spends less time on investigating the worst alternatives. The UCT function in selection should bias computational resources towards the most promising candidates. While the utility of a node for action selection is the size of the reachability set, we chose to divide this value by the number of visits for the selection phase. We thus keep the value between 0.0 and 1.0 and the nodes remain comparable. If this was not the case, nodes that had been explored more would be preferred as the number of found reachable states heavily depends on the number of visits.

Before we evaluate this approach, referred to as UCT from now on, we introduce three modifications to the algorithm.

\subsection{UCT with Novelty Bias}

For this modification we check every time a child node adds a reachable state to its parents reachability set, if this state is new to the set. If this is the case, no other child of that parent has an expanded action sequence leading to that sensor state yet. In case of a novel addition, the child increases its novelty counter by one (\textcolor{blue}{line 29} in Alg. 1). 

The novelty counter is added to the size of the reachability set in the UCT function. This means that the performance of the nodes is now not only defined by how many different states it found, but also by how many of those states were novel contribution to the reachability set of its parent node. 

\textbf{Rationale:} This modification is aimed at biasing exploration towards sub-sequences that add novel states. Consider an agent in front of a doorway. One action leads through it, while the others keep the agent moving around in the room. The sequences that stay in the room end up overlapping, and while they might each lead to a sizeable amount of states, they all lead to the same states. In contrast, sequences that go through the door can add new reachable states to the parent nodes, and should therefore be preferred.  

\subsection{Aggregated Empowerment}

For aggregated empowerment, which we will refer to as UCTa, we not only consider the sensor states reached at tree depth $n$, but we also consider all reachable states along the way. This is achieved by simply adding the sensor state of the current world state to every newly expanded node and propagating it upward (\textcolor{red}{line 18} in Alg.1). 

\textbf{Rationale:} Aggregated empowerment corresponds to a somewhat different quantity, and the implications of this measure go beyond the scope of this paper. Using this value to approximate regular empowerment however still has the advantages that it allows to differentiate between single action sequences. Normally, each sequence reaches exactly one state. But with aggregated empowerment, sequences that go through different sensor sequences along the way are considered better. This difference can indicate that the agent is still operational and able to affect the world.

More conceptually, it also allows to differentiate between different sequences, even if they ultimately end in the same state, e.g. death. In this case, the agents prefers to live a less boring life. In a way, this introduces a form of count-based novelty into the empowerment calculation.    

\subsection{UCT with Full Branching}

For $n$-step empowerment with $k$-step \textit{full branching}, the algorithm only expands the tree to a depth of $n-k$, i.e. it stops $k$ steps before it reaches the time horizon. It then fully expands the tree from that node for the remaining $k$ steps with depth-first search, and eventually propagates all found sensor states upwards. All full branching examples in this paper use $1$-step full branching. This modification is implemented by the extension of the \textbf{BRANCH} function in Alg.1, highlighted in \textcolor{green}{green}.

\subsubsection{Rationale} Full branching also aims to differentiate between sequences. It basically computes 1-step deterministic empowerment for the semi-leaf node. This local empowerment ideally gives us an idea of how empowered close-by states are, and thus should guide our exploration towards- or away from action sequences starting with the same actions.

\section{Evaluation}

\subsection{Simulation Model}

To evaluate the different UCT algorithms we implemented a three-dimensional block world, similar to \cite{Salge2014a}. The world is a three dimensional grid, and each grid cell is empty or contains exactly one block such as earth, the agent and lava. The agent can try to move in the four cardinal directions (\textit{north, east, west, south}). The move will be successful, if a) the target location is empty, or if b) the target location is filled, but the one above is empty. In the second case, the agent will `climb' to the location above the target block. In all other cases the move fails (is blocked) and the agent remains in its current position. The agent can also \emph{act} in all six directions (\textit{above, below, north, south, east, west}). This action is context-sensitive on the agent's inventory, which can hold exactly one block. If the inventory is empty, the agent will try to take the block from the target location into its inventory, if there is any. If the inventory contains a block, the agent will try to place it. This will succeed if the target location is empty. If an action fails, the world remains unchanged. There are two additional actions, \emph{waiting} and \emph{destroying the current block} in the inventory. Overall, this gives the agent $12$ actions in each time step.

Between actions the world simulates liquid flow and gravity. Agents and lava are affected by gravity, i.e. they fall until they rest above a filled block. Earth blocks are not affected by gravity. Lava spreads to neighbouring tiles every $4$ time steps, if they are empty. Lava is an environmental hazard, and the agent dies if it is next to a lava block. `Death' in this case means that the agent's actions have no effect any more on its world. The sensors considered for the empowerment computation capture the agent's $x,y,z$ position. 

\subsection{Method}

\begin{figure}[htp]
\centering
\begin{tikzpicture}[scale=0.6]
\cube{0}{4}{0}{20}{50}{70}
\cube{0}{1}{1}{23}{53}{73}
\cube{0}{4}{1}{23}{53}{73}
\cube{0}{1}{2}{26}{56}{76}
\cube{0}{4}{2}{26}{56}{76}
\cube{0}{0}{3}{29}{59}{79}
\cube{0}{2}{3}{29}{59}{79}
\cube{0}{5}{3}{29}{59}{79}
\cube{0}{1}{4}{32}{62}{82}
\cube{0}{2}{4}{32}{62}{82}
\cube{0}{3}{4}{32}{62}{82}
\cube{0}{4}{4}{32}{62}{82}
\cube{0}{5}{4}{32}{62}{82}
\cube{0}{2}{5}{35}{65}{85}
\cube{0}{4}{5}{35}{65}{85}
\cube{0}{5}{5}{35}{65}{85}
\cube{1}{0}{0}{20}{50}{70}
\cube{1}{2}{0}{20}{50}{70}
\cube{1}{0}{1}{23}{53}{73}
\cube{1}{1}{1}{23}{53}{73}
\cubeLava{1}{2}{1}
\cube{1}{3}{1}{23}{53}{73}
\cube{1}{4}{1}{23}{53}{73}
\cube{1}{5}{1}{23}{53}{73}
\cube{1}{0}{2}{26}{56}{76}
\cube{1}{1}{2}{26}{56}{76}
\cube{1}{3}{2}{26}{56}{76}
\cube{1}{4}{2}{26}{56}{76}
\cube{1}{0}{3}{29}{59}{79}
\cube{1}{1}{3}{29}{59}{79}
\cube{1}{2}{3}{29}{59}{79}
\cubeAgent{1}{3}{3}
\cube{1}{5}{3}{29}{59}{79}
\cube{1}{1}{4}{32}{62}{82}
\cube{1}{3}{4}{32}{62}{82}
\cube{1}{4}{4}{32}{62}{82}
\cube{1}{5}{4}{32}{62}{82}
\cube{1}{0}{5}{35}{65}{85}
\cube{1}{4}{5}{35}{65}{85}
\cube{2}{1}{0}{20}{50}{70}
\cube{2}{2}{0}{20}{50}{70}
\cube{2}{3}{0}{20}{50}{70}
\cube{2}{2}{1}{23}{53}{73}
\cube{2}{3}{1}{23}{53}{73}
\cube{2}{1}{2}{26}{56}{76}
\cube{2}{0}{3}{29}{59}{79}
\cube{2}{2}{3}{29}{59}{79}
\cube{2}{0}{4}{32}{62}{82}
\cube{2}{2}{4}{32}{62}{82}
\cube{2}{2}{5}{35}{65}{85}
\cubeLava{2}{4}{5}
\cube{2}{5}{5}{35}{65}{85}
\cube{3}{0}{0}{20}{50}{70}
\cube{3}{0}{1}{23}{53}{73}
\cube{3}{4}{1}{23}{53}{73}
\cube{3}{5}{1}{23}{53}{73}
\cube{3}{0}{2}{26}{56}{76}
\cube{3}{4}{2}{26}{56}{76}
\cube{3}{1}{4}{32}{62}{82}
\cube{3}{1}{5}{35}{65}{85}
\cube{3}{2}{5}{35}{65}{85}
\cube{4}{3}{0}{20}{50}{70}
\cube{4}{1}{1}{23}{53}{73}
\cube{4}{3}{1}{23}{53}{73}
\cube{4}{4}{1}{23}{53}{73}
\cube{4}{0}{2}{26}{56}{76}
\cube{4}{1}{2}{26}{56}{76}
\cube{4}{2}{2}{26}{56}{76}
\cube{4}{0}{3}{29}{59}{79}
\cube{4}{4}{3}{29}{59}{79}
\cube{4}{3}{4}{32}{62}{82}
\cube{4}{3}{5}{35}{65}{85}
\cube{4}{4}{5}{35}{65}{85}
\cube{4}{5}{5}{35}{65}{85}
\cube{5}{0}{0}{20}{50}{70}
\cube{5}{3}{0}{20}{50}{70}
\cube{5}{4}{0}{20}{50}{70}
\cube{5}{2}{1}{23}{53}{73}
\cube{5}{3}{1}{23}{53}{73}
\cube{5}{4}{1}{23}{53}{73}
\cube{5}{5}{1}{23}{53}{73}
\cube{5}{1}{2}{26}{56}{76}
\cube{5}{1}{4}{32}{62}{82}
\cube{5}{0}{5}{35}{65}{85}
\cube{5}{1}{5}{35}{65}{85}
\cube{5}{5}{5}{35}{65}{85}
\end{tikzpicture}
\caption{Typical randomly generated world used for the quantitative evaluation. Two red lava blocks are visible among the grey earth blocks. The agent is colored blue.}
\label{fig:randomWorld}
\end{figure}
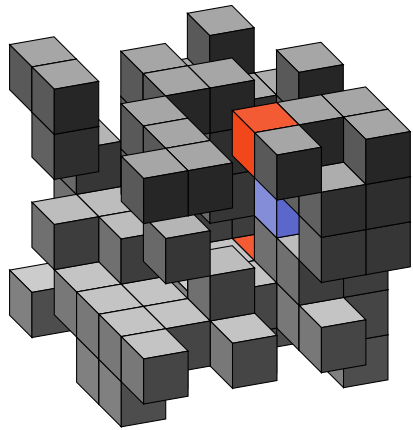

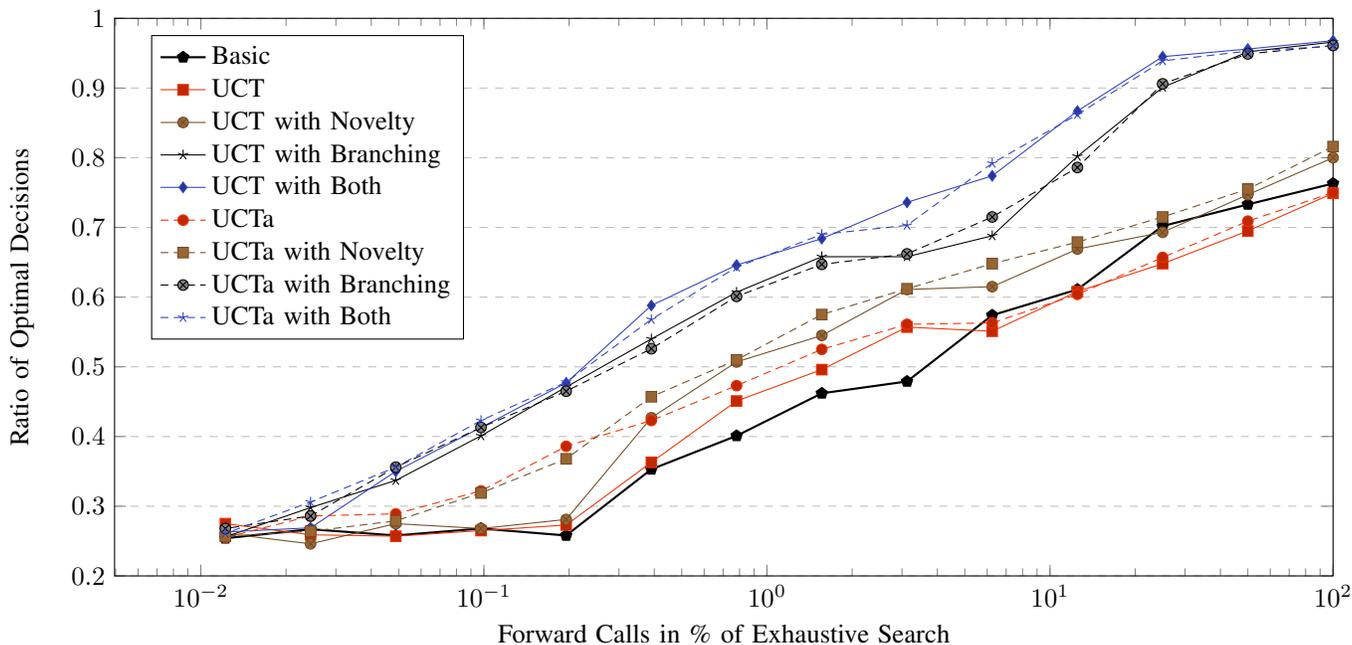
\begin{figure*}[htb]
\begin{tikzpicture}[scale=1]
\begin{semilogxaxis}[
    title={},
    width= \linewidth,
	height=9cm,
    xlabel={Forward Calls in \% of Exhaustive Search},
    ylabel={Ratio of Optimal Decisions},
    xmin=0, xmax=100,
    ymin=0.2, ymax=1,
    legend pos= north west,
    legend cell align=left,
    ymajorgrids=true,
    grid style=dashed,
]
\addlegendentry{Basic}
\addplot[mark=pentagon*,thick] table[x index=0,y index=1,col sep=comma]
{resultsperformance0.csv};
\label{graph:basic}

\addlegendentry{UCT}
\addplot table[x index=0,y index=2,col sep=comma] {resultsperformance0.csv};
\label{graph:UCT}

\addlegendentry{UCT with Novelty}
\addplot table[x index=0,y index=3,col sep=comma] {resultsperformance0.csv};
\label{graph:UCTNovelty}

\addlegendentry{UCT with Branching}
\addplot table[x index=0,y index=4,col sep=comma] {resultsperformance0.csv};
\label{graph:UCTBranching}
\addlegendentry{UCT with Both}
\addplot table[x index=0,y index=5,col sep=comma] {resultsperformance0.csv};
\label{graph:UCTBoth}
\addlegendentry{UCTa}
\addplot table[x index=0,y index=6,col sep=comma] {resultsperformance0.csv};
\label{graph:UCTa}
\addlegendentry{UCTa with Novelty}
\addplot table[x index=0,y index=7,col sep=comma] {resultsperformance0.csv};
\label{graph:UCTaNovelty}
\addlegendentry{UCTa with Branching}
\addplot table[x index=0,y index=8,col sep=comma] {resultsperformance0.csv};
\label{graph:UCTaBranching}
\addlegendentry{UCTa with Both}
\addplot table[x index=0,y index=9,col sep=comma] {resultsperformance0.csv};
\label{graph:UCTaBoth}
\end{semilogxaxis}
\end{tikzpicture}
\caption{Ratio of selecting the optimal action for different budgets of forward model calls. The evaluation is based on 1000 different random worlds, computing 4-step action sequences.}
\label{fig:optdec}
\end{figure*}

\begin{figure*}[htb]
\begin{tikzpicture}[scale=1]
\begin{semilogxaxis}[
    title={},
    width= \linewidth,
	height=9cm,
    xlabel={Forward Calls in \% of Exhaustive Search},
    ylabel={Average Relative Performance},
    xmin=0, xmax=100,
    ymin=0.65, ymax=1,
    legend pos= south east,
    legend cell align=left,
    ymajorgrids=true,
    grid style=dashed,
]

\addlegendentry{Basic}
\addplot[mark=pentagon*,thick] table[x index=0,y index=1,col sep=comma]
{resultsperformance.csv};
\addlegendentry{UCT}
\addplot table[x index=0,y index=2,col sep=comma] {resultsperformance.csv};
\addlegendentry{UCT with Novelty}
\addplot table[x index=0,y index=3,col sep=comma] {resultsperformance.csv};
\addlegendentry{UCT with Branching}
\addplot table[x index=0,y index=4,col sep=comma] {resultsperformance.csv};
\addlegendentry{UCT with Both}
\addplot table[x index=0,y index=5,col sep=comma] {resultsperformance.csv};
\addlegendentry{UCTa}
\addplot table[x index=0,y index=6,col sep=comma] {resultsperformance.csv};
\addlegendentry{UCTa with Novelty}
\addplot table[x index=0,y index=7,col sep=comma] {resultsperformance.csv};
\addlegendentry{UCTa with Branching}
\addplot table[x index=0,y index=8,col sep=comma] {resultsperformance.csv};
\addlegendentry{UCTa with Both}
\addplot table[x index=0,y index=9,col sep=comma] {resultsperformance.csv};
\end{semilogxaxis}
\end{tikzpicture}
\caption{Average relative performance (in reachable states) for the picked actions compared to the optimal action, for different budgets of forward model calls. The evaluation is based on 1000 different random worlds, computing 4-step action sequences.}
\label{fig:perf}
\end{figure*}
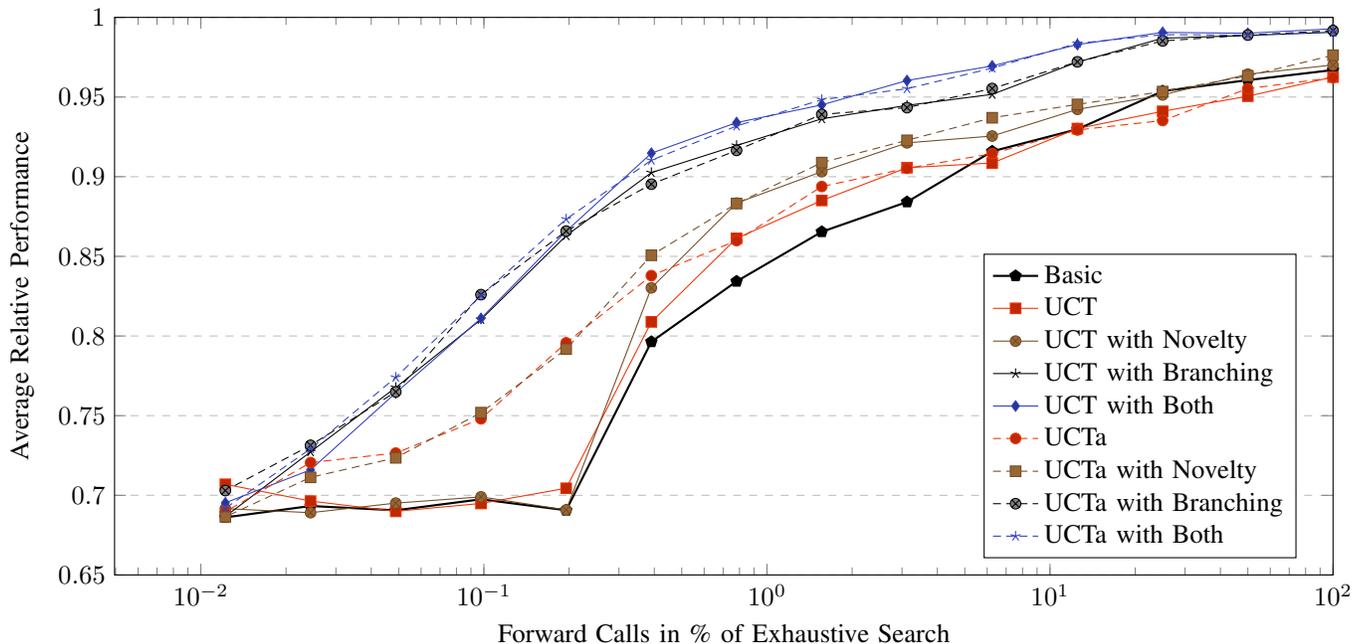 

For the quantitative evaluation we created $1000$ different random worlds of size 7x7x7. Each block has a 40\% chance to be an earth block, a 2\% chance to be a lava block, and remains empty otherwise. Fig.~\ref{fig:randomWorld} shows an example world. The agent (blue) is placed in a random position, replacing the block in that position if necessary.

For each world we computed $4$-step empowerment with an exhaustive depth-first search to obtain a baseline comparison. We recorded the number of reachable states, which determine the empowerment, for each immediate successor state and the action leading to that state from the root. This gives us an empowerment value for each action. The UCT algorithm can ultimately be applied to longer action sequences, but we chose a horizon of 4 for the quantitative evaluation, as it allows us to compare the approximations to an exact baseline. 

We evaluated 9 different algorithms. The \emph{basic} agent realises just random sampling, as describe in \cite{Salge2014a}. We evaluated the described \emph{UCT} algorithm with each of the three modifications (\emph{novelty bias}, \emph{aggregated empowerment}, \emph{full branching} for $1$ step) turned on or off. Additionally, we evaluated the \emph{UCTa} algorithm based on aggregated empowerment for each modification. 

For each of the 1000 different worlds, each of the 9 algorithms tried to find the most empowered action. Our goal was to see how the algorithm's performance would degrade if they were given less computational resources. To have a hardware-independent comparison, we limited each algorithm to a certain number of forward model calls, which are used every time a world is advanced by applying an action. This decision was based on a preliminary analysis, showing that forward model calls were (unsurprisingly) the main contributor to computational load. We recorded the number of forward model calls used by the depth-first algorithm from the exhaustive baseline computation. This number, 22621, was considered as 100 \% of needed calls. We then evaluated all 9 algorithms on all 1000 worlds by giving them only $1/2$, $1/4$, $1/8$, etc, of the forward calls required for the baseline.

The performance of returned actions was evaluated in two ways. We checked, in comparison to the baseline computation, if the action selected by an action is optimal, i.e. if it has as many reachable states as the best answer. The average performance in this case is the ratio of how many of the chosen actions where optimal. We also implemented a second performance measure, where we compared the empowerment of the given action to the empowerment of the best action. If for example the best action would lead to a state with 20 reachable states, and the given answer would lead to a state with 10 reachable states, the performance would be 0.5. The performance is then averaged over the 1000 tested worlds. 

\subsection{Results}

Figs.~\ref{fig:optdec} and \ref{fig:perf} show the results of the evaluation. As expected, both performance measures get worse if the agents get fewer forward calls to determine the best action. The important result here is that the UCT algorithms, especially those with modifications, outperform the basic agent from \cite{Salge2014a} for less forward calls. 

If we look at the graph in more detail, we see that for the least amount of samples, all agents perform at about $0.7$ in Fig.~\ref{fig:perf} and at about 0.25 in Fig.~\ref{fig:optdec}. This is basically the performance of a random agent that occasionally finds the best action by chance, or picks an action that has decent empowerment. Keep in mind that in some worlds several actions are optimal, or close to optimal. 

The remaining graph can be split into three parts. For the first 5 sample points, i.e. the part with the least forward calls, we see that the basic agent \ref{graph:basic}, as well as UCT \ref{graph:UCT} and UCT with novelty \ref{graph:UCTNovelty} all perform at the random level. At this point so few forward calls are made that each child of the root is only expanded once, i.e. has one full actions sequence going through it evaluated. This does not allow yet for an informed choice, as all sequences in this case lead to exactly one sensor state, making all choices equivalent. The algorithms with aggregated empowerment (\ref{graph:UCTa},\ref{graph:UCTaNovelty},\ref{graph:UCTaBranching},\ref{graph:UCTaBoth}) perform better here, as even a single sequence becomes informative. Each sequence might be visiting more or fewer different states, giving a better picture of how much an agent that picked a specific first action can affect the world afterwards. If, for example, the first action would lead the agent to touch lava and die, or fall in a hole, then all successive states of that sequence would be identical. The four algorithms with full 1-step branching (\ref{graph:UCTBranching},\ref{graph:UCTBoth},\ref{graph:UCTaBranching},\ref{graph:UCTaBoth}) perform best in the segment, as they create the most information for a single sequence. By evaluating how many states could be reached from the second to last step, the algorithms can differentiate between different starting actions. These approaches cost slightly more, as they use 11 additional forward calls in the end, but then saves forward calls by not having to go down the same full sequence several times. 

In the middle segment of the graph we see an increase in performance for all algorithms. Noteworthy here is that for every configuration the variant with novelty outperforms the respective variant without novelty bias. As we now have enough samples so that children are fully expanded, the UCT selection comes into play. The bias towards those states that brought novel contributions seems to guide us towards better states. Aggregated Empowerment seems to have little effect on performance in this segment. 

Towards the end, where the algorithms get nearly enough calls to exhaustively sample the space we can see that the marginal difference in performance gets slimmer as the performances increase overall. The algorithms basically split into two groups, those with full 1-step branching (\ref{graph:UCTBranching},\ref{graph:UCTBoth},\ref{graph:UCTaBranching},\ref{graph:UCTaBoth}) clearly outperforming those without. We should also note that the basic agent performing random sampling gets more competitive once we get to nearly \mbox{100\%} of the needed samples. 

\subsection{Bridge Example}

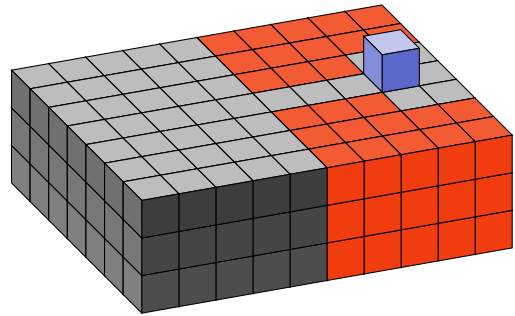
\begin{figure}[htp]
\centering
\begin{tikzpicture}[scale=0.5]

\cubeLava{0}{0}{0}
\cubeLava{0}{1}{0}
\cubeLava{0}{2}{0}
\cubeLava{0}{3}{0}
\cubeLava{0}{4}{0}
\cubeLava{0}{5}{0}
\cubeLava{0}{6}{0}
\cubeLava{0}{0}{1}
\cubeLava{0}{1}{1}
\cubeLava{0}{2}{1}
\cubeLava{0}{3}{1}
\cubeLava{0}{4}{1}
\cubeLava{0}{5}{1}
\cubeLava{0}{6}{1}
\cubeLava{0}{0}{2}
\cubeLava{0}{1}{2}
\cube{0}{2}{2}{26}{56}{76}
\cube{0}{3}{2}{26}{56}{76}
\cube{0}{4}{2}{26}{56}{76}
\cubeLava{0}{5}{2}
\cubeLava{0}{6}{2}
\cubeLava{1}{0}{0}
\cubeLava{1}{1}{0}
\cubeLava{1}{2}{0}
\cubeLava{1}{3}{0}
\cubeLava{1}{4}{0}
\cubeLava{1}{5}{0}
\cubeLava{1}{6}{0}
\cubeLava{1}{0}{1}
\cubeLava{1}{1}{1}
\cubeLava{1}{2}{1}
\cubeLava{1}{3}{1}
\cubeLava{1}{4}{1}
\cubeLava{1}{5}{1}
\cubeLava{1}{6}{1}
\cubeLava{1}{0}{2}
\cubeLava{1}{1}{2}
\cube{1}{2}{2}{26}{56}{76}
\cube{1}{3}{2}{26}{56}{76}
\cube{1}{4}{2}{26}{56}{76}
\cubeLava{1}{5}{2}
\cubeLava{1}{6}{2}
\cubeAgent{1}{3}{3}
\cubeLava{2}{0}{0}
\cubeLava{2}{1}{0}
\cubeLava{2}{2}{0}
\cubeLava{2}{3}{0}
\cubeLava{2}{4}{0}
\cubeLava{2}{5}{0}
\cubeLava{2}{6}{0}
\cubeLava{2}{0}{1}
\cubeLava{2}{1}{1}
\cubeLava{2}{2}{1}
\cubeLava{2}{3}{1}
\cubeLava{2}{4}{1}
\cubeLava{2}{5}{1}
\cubeLava{2}{6}{1}
\cubeLava{2}{0}{2}
\cubeLava{2}{1}{2}
\cubeLava{2}{2}{2}
\cube{2}{3}{2}{26}{56}{76}
\cubeLava{2}{4}{2}
\cubeLava{2}{5}{2}
\cubeLava{2}{6}{2}
\cubeLava{3}{0}{0}
\cubeLava{3}{1}{0}
\cubeLava{3}{2}{0}
\cubeLava{3}{3}{0}
\cubeLava{3}{4}{0}
\cubeLava{3}{5}{0}
\cubeLava{3}{6}{0}
\cubeLava{3}{0}{1}
\cubeLava{3}{1}{1}
\cubeLava{3}{2}{1}
\cubeLava{3}{3}{1}
\cubeLava{3}{4}{1}
\cubeLava{3}{5}{1}
\cubeLava{3}{6}{1}
\cubeLava{3}{0}{2}
\cubeLava{3}{1}{2}
\cubeLava{3}{2}{2}
\cube{3}{3}{2}{26}{56}{76}
\cubeLava{3}{4}{2}
\cubeLava{3}{5}{2}
\cubeLava{3}{6}{2}
\cubeLava{4}{0}{0}
\cubeLava{4}{1}{0}
\cubeLava{4}{2}{0}
\cubeLava{4}{3}{0}
\cubeLava{4}{4}{0}
\cubeLava{4}{5}{0}
\cubeLava{4}{6}{0}
\cubeLava{4}{0}{1}
\cubeLava{4}{1}{1}
\cubeLava{4}{2}{1}
\cubeLava{4}{3}{1}
\cubeLava{4}{4}{1}
\cubeLava{4}{5}{1}
\cubeLava{4}{6}{1}
\cubeLava{4}{0}{2}
\cubeLava{4}{1}{2}
\cubeLava{4}{2}{2}
\cube{4}{3}{2}{26}{56}{76}
\cubeLava{4}{4}{2}
\cubeLava{4}{5}{2}
\cubeLava{4}{6}{2}
\cube{5}{0}{0}{20}{50}{70}
\cube{5}{1}{0}{20}{50}{70}
\cube{5}{2}{0}{20}{50}{70}
\cube{5}{3}{0}{20}{50}{70}
\cube{5}{4}{0}{20}{50}{70}
\cube{5}{5}{0}{20}{50}{70}
\cube{5}{6}{0}{20}{50}{70}
\cube{5}{0}{1}{23}{53}{73}
\cube{5}{1}{1}{23}{53}{73}
\cube{5}{2}{1}{23}{53}{73}
\cube{5}{3}{1}{23}{53}{73}
\cube{5}{4}{1}{23}{53}{73}
\cube{5}{5}{1}{23}{53}{73}
\cube{5}{6}{1}{23}{53}{73}
\cube{5}{0}{2}{26}{56}{76}
\cube{5}{1}{2}{26}{56}{76}
\cube{5}{2}{2}{26}{56}{76}
\cube{5}{3}{2}{26}{56}{76}
\cube{5}{4}{2}{26}{56}{76}
\cube{5}{5}{2}{26}{56}{76}
\cube{5}{6}{2}{26}{56}{76}
\cube{6}{0}{0}{20}{50}{70}
\cube{6}{1}{0}{20}{50}{70}
\cube{6}{2}{0}{20}{50}{70}
\cube{6}{3}{0}{20}{50}{70}
\cube{6}{4}{0}{20}{50}{70}
\cube{6}{5}{0}{20}{50}{70}
\cube{6}{6}{0}{20}{50}{70}
\cube{6}{0}{1}{23}{53}{73}
\cube{6}{1}{1}{23}{53}{73}
\cube{6}{2}{1}{23}{53}{73}
\cube{6}{3}{1}{23}{53}{73}
\cube{6}{4}{1}{23}{53}{73}
\cube{6}{5}{1}{23}{53}{73}
\cube{6}{6}{1}{23}{53}{73}
\cube{6}{0}{2}{26}{56}{76}
\cube{6}{1}{2}{26}{56}{76}
\cube{6}{2}{2}{26}{56}{76}
\cube{6}{3}{2}{26}{56}{76}
\cube{6}{4}{2}{26}{56}{76}
\cube{6}{5}{2}{26}{56}{76}
\cube{6}{6}{2}{26}{56}{76}
\cube{7}{0}{0}{20}{50}{70}
\cube{7}{1}{0}{20}{50}{70}
\cube{7}{2}{0}{20}{50}{70}
\cube{7}{3}{0}{20}{50}{70}
\cube{7}{4}{0}{20}{50}{70}
\cube{7}{5}{0}{20}{50}{70}
\cube{7}{6}{0}{20}{50}{70}
\cube{7}{0}{1}{23}{53}{73}
\cube{7}{1}{1}{23}{53}{73}
\cube{7}{2}{1}{23}{53}{73}
\cube{7}{3}{1}{23}{53}{73}
\cube{7}{4}{1}{23}{53}{73}
\cube{7}{5}{1}{23}{53}{73}
\cube{7}{6}{1}{23}{53}{73}
\cube{7}{0}{2}{26}{56}{76}
\cube{7}{1}{2}{26}{56}{76}
\cube{7}{2}{2}{26}{56}{76}
\cube{7}{3}{2}{26}{56}{76}
\cube{7}{4}{2}{26}{56}{76}
\cube{7}{5}{2}{26}{56}{76}
\cube{7}{6}{2}{26}{56}{76}
\cube{8}{0}{0}{20}{50}{70}
\cube{8}{1}{0}{20}{50}{70}
\cube{8}{2}{0}{20}{50}{70}
\cube{8}{3}{0}{20}{50}{70}
\cube{8}{4}{0}{20}{50}{70}
\cube{8}{5}{0}{20}{50}{70}
\cube{8}{6}{0}{20}{50}{70}
\cube{8}{0}{1}{23}{53}{73}
\cube{8}{1}{1}{23}{53}{73}
\cube{8}{2}{1}{23}{53}{73}
\cube{8}{3}{1}{23}{53}{73}
\cube{8}{4}{1}{23}{53}{73}
\cube{8}{5}{1}{23}{53}{73}
\cube{8}{6}{1}{23}{53}{73}
\cube{8}{0}{2}{26}{56}{76}
\cube{8}{1}{2}{26}{56}{76}
\cube{8}{2}{2}{26}{56}{76}
\cube{8}{3}{2}{26}{56}{76}
\cube{8}{4}{2}{26}{56}{76}
\cube{8}{5}{2}{26}{56}{76}
\cube{8}{6}{2}{26}{56}{76}
\cube{9}{0}{0}{20}{50}{70}
\cube{9}{1}{0}{20}{50}{70}
\cube{9}{2}{0}{20}{50}{70}
\cube{9}{3}{0}{20}{50}{70}
\cube{9}{4}{0}{20}{50}{70}
\cube{9}{5}{0}{20}{50}{70}
\cube{9}{6}{0}{20}{50}{70}
\cube{9}{0}{1}{23}{53}{73}
\cube{9}{1}{1}{23}{53}{73}
\cube{9}{2}{1}{23}{53}{73}
\cube{9}{3}{1}{23}{53}{73}
\cube{9}{4}{1}{23}{53}{73}
\cube{9}{5}{1}{23}{53}{73}
\cube{9}{6}{1}{23}{53}{73}
\cube{9}{0}{2}{26}{56}{76}
\cube{9}{1}{2}{26}{56}{76}
\cube{9}{2}{2}{26}{56}{76}
\cube{9}{3}{2}{26}{56}{76}
\cube{9}{4}{2}{26}{56}{76}
\cube{9}{5}{2}{26}{56}{76}
\cube{9}{6}{2}{26}{56}{76}
\end{tikzpicture}
\caption{Bridge example world. The agent (blue) should cross over the narrow bridge and avoid the lava (red).}
\label{fig:bridge}
\end{figure}

The qualitative effect of the different algorithms becomes more evident if we are looking at a concrete example. In Fig.~\ref{fig:bridge} we see an agent on a small platform surrounded by lava. The agent should cross over the narrow bridge to reach the much larger area. This would massively increase its empowerment. Looking at 10-step empowerment, the agent has a long enough time horizon to figure this out. But it is infeasible to evaluate all 61917364224 10-step action sequences, so we do not have a ground truth to evaluate against. This is the kind of scenario the algorithm in this paper was developed for. 

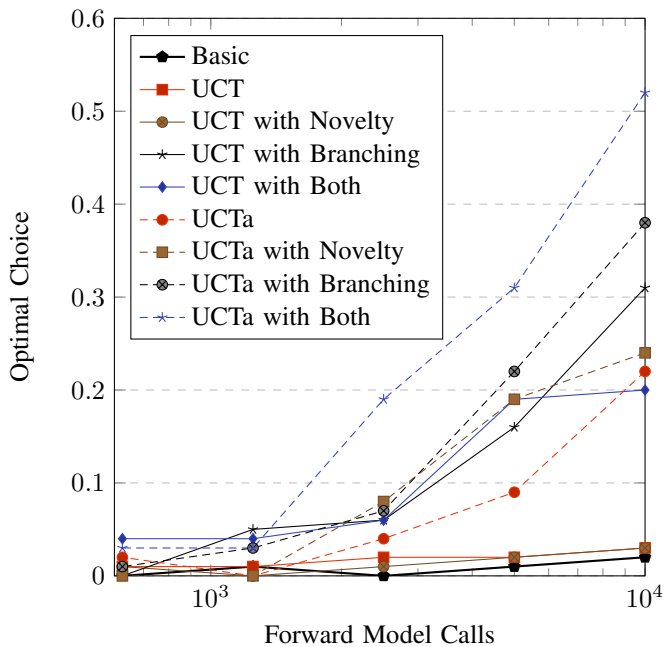
\begin{figure}[htb]
\begin{tikzpicture}[scale=1]
\begin{semilogxaxis}[
    title={},
    width= \linewidth,
	height=9cm,
    xlabel={Forward Model Calls},
    ylabel={Optimal Choice},
    xmin=600, xmax=10001,
    ymin=0.0, ymax=0.6,
    legend pos= north west,
    legend cell align=left,
    ymajorgrids=true,
    grid style=dashed,
]
\addlegendentry{Basic}
\addplot[mark=pentagon*,thick] table[x index=0,y index=1,col sep=comma]
{bridge.csv};
\addlegendentry{UCT}
\addplot table[x index=0,y index=2,col sep=comma] {bridge.csv};
\addlegendentry{UCT with Novelty}
\addplot table[x index=0,y index=3,col sep=comma] {bridge.csv};
\addlegendentry{UCT with Branching}
\addplot table[x index=0,y index=4,col sep=comma] {bridge.csv};
\addlegendentry{UCT with Both}
\addplot table[x index=0,y index=5,col sep=comma] {bridge.csv};
\addlegendentry{UCTa}
\addplot table[x index=0,y index=6,col sep=comma] {bridge.csv};
\addlegendentry{UCTa with Novelty}
\addplot table[x index=0,y index=7,col sep=comma] {bridge.csv};
\addlegendentry{UCTa with Branching}
\addplot table[x index=0,y index=8,col sep=comma] {bridge.csv};
\addlegendentry{UCTa with Both}
\addplot table[x index=0,y index=9,col sep=comma] {bridge.csv};
\end{semilogxaxis}
\end{tikzpicture}
\caption{Evaluation of the bridge example in Fig.~\ref{fig:bridge} for 10-step sequences. The optimal choice is going towards the bridge.}
\label{fig:bridgegraph}
\end{figure}

We evaluated this example with 10000 forward calls and less, for all previously described algorithmic variations. We ran the world seen in Fig.~\ref{fig:bridge} 100 times, and we checked how often the agent started moving onto the bridge leading to more empowerment. From our conceptual understanding of discrete empowerment we know that this is the most empowered option. The graph in Fig.~\ref{fig:bridgegraph} shows the results of the evaluation. The basic \ref{graph:basic} and the UCT agent \ref{graph:UCT} struggle to find the optimal action, as fewer than 5\% actually move towards the bridge with the first action. The bridge defines a bottleneck that first has to be crossed with an initial, very specific 4-step sequence, and biasing the exploration of the graph towards this one sequence does not happen in this case. The addition of novelty bias seems to add relatively little, but full branching and aggregated empowerment pick the optimal action far more often. The algorithm that combines all modifications performs best, finding the optimal solution 52\% of the time compared to 2\% for the basic agent from \cite{Salge2014a}. As a side note, this is also an example of how directed behaviour arises in a world without the kind of utility function typical to most games. The agent prefers to be on the bigger platform, as it can move around more and dig down to obtain blocks to climb up. 

\section{conclusion}

The results indicate that all three modifications introduced in this paper improve the performance of empowerment maximising agents that have a limited amount of computational resources. While the approach is intended to be used for longer, i.e. above 10-steps, action sequences, we evaluated 4-step sequences. Due to computational intractability, we do not have a baseline comparison for longer sequences. The bridge example is promising though, as it showed that the modified algorithms can deal with bottlenecks in longer sequences, something the basic algorithm from previous work \cite{Salge2014a} struggled with considerably. 

\subsection{Future Work}

This technical improvement opens up possible applications of empowerment in the discrete and deterministic domain. For example, this would make it more feasible to apply an empowerment-biased agent to games such as those found in general game-playing competitions \cite{genesereth2005general,perez2016general}. The other possible extension of this work is to extend it to non-deterministic but discrete models. While there are faster approximations for this domain \cite{Anthony2014}, it might prove useful to apply the extensive catalogue of MCTS enhancement to this problem. 

\section*{Acknowledgement}
\small
CS is funded by the EU Horizon 2020 programme under the Marie Sklodowska-Curie grant 705643. CG is funded by EPSRC grant [EP/L015846/1] (IGGI).
RC gratefully acknowledges the financial support from Honda Research Institute Europe (HRI-EU).

\small
\bibliographystyle{IEEEtran}
\bibliography{references}

\end{document}